\documentclass[10pt,journal,compsoc]{IEEEtran}

\usepackage{listings}

\usepackage{epsfig}
\usepackage[draft]{pgf}

\ifCLASSOPTIONcompsoc
  \usepackage[nocompress]{cite}
\else
  \usepackage{cite}
\fi
\ifCLASSINFOpdf
\else
\fi

\newcommand{\TS}{TS}
\newcommand{\TSlong}{Time Series}
\newcommand{\CC}{CC}
\newcommand{\WFlong}{workflow}
\newcommand{\DM}{DM}
\newcommand{\ML}{ML}
\newcommand{\IT}{IT}

\newcommand{\ARIMA}{ARIMA}
\newcommand{\SVM}{SVM}

\newcommand{\pacf}{PACF}
\newcommand{\acf}{ACF}
\newcommand{\stl}{STL}
\newcommand{\tswfschema}{\textit{tswf-schema}}
\newcommand{\ETLlong}{extract,transform,load}
\newcommand{\DLlong}{Deep Learning}
\newcommand{\NNlong}{Neural Networks}
\newcommand{\SVMlong}{Support Vector Machine}
\newcommand{\RFlong}{Random Forest}
\newcommand{\R}{\textit{R}}
\newcommand{\jsonld}{\textit{json-ld}}
\newcommand{\CQs}{CQs}
\newcommand{\ttl}{ttl}
\newcommand{\marketplace}{Marketplace}
\newcommand{\python}{Python}
\newcommand{\dmccschema}{\texttt{dmcc-schema}}

\usepackage{xcolor}
\usepackage{listings}
\colorlet{punct}{red!60!black}
\definecolor{background}{HTML}{EEEEEE}
\definecolor{delim}{RGB}{20,105,176}
\colorlet{numb}{magenta!60!black}

\lstdefinelanguage{json}{
    basicstyle=\normalfont\ttfamily,
    numbers=right,
    numberstyle=\scriptsize,
    stepnumber=1,
    numbersep=8pt,
    showstringspaces=false,
    breaklines=true,
    frame=lines,
    backgroundcolor=\color{background},
    literate=
     *{0}{{{\color{numb}0}}}{1}
      {1}{{{\color{numb}1}}}{1}
      {2}{{{\color{numb}2}}}{1}
      {3}{{{\color{numb}3}}}{1}
      {4}{{{\color{numb}4}}}{1}
      {5}{{{\color{numb}5}}}{1}
      {6}{{{\color{numb}6}}}{1}
      {7}{{{\color{numb}7}}}{1}
      {8}{{{\color{numb}8}}}{1}
      {9}{{{\color{numb}9}}}{1}
      {:}{{{\color{punct}{:}}}}{1}
      {,}{{{\color{punct}{,}}}}{1}
      {\{}{{{\color{delim}{\{}}}}{1}
      {\}}{{{\color{delim}{\}}}}}{1}
      {[}{{{\color{delim}{[}}}}{1}
      {]}{{{\color{delim}{]}}}}{1},
}

\begin{document}
%
\title{Semantic of Cloud Computing services for Time Series workflow}

%
%
%
%

\author{Manuel~Parra,
        Francisco J.~Bald\'{a}n,
        Ghislain~Atemezing 
        and~Jos\'{e} M.~Ben\'{i}tez
\IEEEcompsocitemizethanks{\IEEEcompsocthanksitem Manuel Parra-Roy\'{o}n, Francisco J. Bald\'{a}n and Jos\'{e} M. Ben\'{i}tez are with the Department of Computer Sciences and Artificial Intelligence of the University of Granada, Spain.\protect\\
E-mail: {manuelparra,fjbaldan,j.m.benitez}@decsai.ugr.es
\IEEEcompsocthanksitem Ghislain Atemezing is the head of R\&D at Mondeca, Paris, France.}
\thanks{Manuscript received April 19, 2005; revised August 26, 2015.}}

%
%

\markboth{Journal of \LaTeX\ Class Files,~Vol.~14, No.~8, August~2015}%
{Shell \MakeLowercase{\textit{et al.}}: Bare Advanced Demo of IEEEtran.cls for IEEE Computer Society Journals}
%



\IEEEtitleabstractindextext{%
\begin{abstract}


%

Time series (\TS{}) are present in many fields of knowledge, research, and engineering. The processing and analysis of \TS{} are essential in order to extract knowledge from the data and to tackle forecasting or predictive maintenance tasks among others The modeling of \TS{} is a challenging task, requiring high statistical expertise as well as outstanding knowledge about the application of Data Mining (\DM{}) and Machine Learning (\ML{}) methods. The overall work with \TS{} is not limited to the linear application of several techniques, but is composed of an open \WFlong{} of methods and tests.
These \WFlong{}, developed mainly on programming languages, are complicated to execute and run effectively on different systems, including Cloud Computing (\CC{}) environments. The adoption of \CC{} can facilitate the integration and portability of services allowing to adopt solutions towards services Internet Technologies (\IT{}) industrialization. The definition and description of \WFlong{} services for \TS{} open  up a new set of possibilities regarding the reduction of complexity in the deployment of this type of issues in 
\CC{} environments. In this sense, we have designed an effective proposal based on semantic modeling (or vocabulary)  that provides the full description of  \WFlong{} for \TSlong{} modeling as a \CC{} service. Our proposal includes a broad spectrum of the most extended operations, accommodating any \WFlong{} applied to classification, regression, or clustering problems for \TSlong{}, as well as including evaluation measures, information, tests, or machine learning algorithms among others.


\end{abstract}

}

\maketitle

\IEEEdisplaynontitleabstractindextext

%
\IEEEpeerreviewmaketitle

\ifCLASSOPTIONcompsoc
\IEEEraisesectionheading{\section{Introduction}\label{sec:introduction}}
\else
\section{Introduction}
\label{sec:introduction}
\fi

\IEEEPARstart{F}{orecasting}  of weather conditions, the estimation of the value of stock market shares, or the detection of anomalies on industrial processes among others, are part of the set fields where the \TS{} data analysis plays a basic role to tackle knowledge extraction.


Time series is a sequence or sequences of data spaced out in time; events, activities, or devices continuously generate information that is temporarily logged and stored for real-time or post-processed study. The work with \TS{} is one of the fastest growing at the moment, due to the proliferation of the so-called IoT \cite{gubbi2013internet},  for instance.  Increasingly in interest, the use of mobile devices, autonomous vehicles, modern agriculture, or intelligent machines, will produce a huge amount of information in the coming years and a significant percentage of this information will be in the form of \TS{} data \cite{ericsson2018}. 

At the present time, when \CC{} has practically been integrated in a totally transparent way in our relationship with Information Technologies (IT) and Internet, the activities related to data analysis are incrementally being added to the spectrum of services offered by the \CC{} platforms and providers. Analysis and study of \TS{}  will need to be processed as \CC{} services following the NIST\cite{mell2011nist} recommendations such as flexibility, scalability, portability, and security.


The rise of \CC{} in parallel with the increase in computing  capability has led to the deployment of tools and platforms for data mining and data analysis. Both offer  a wide range of methods, functions and algorithms to perform data processing at all scales, either from the desktop \cite{hall2009weka}, \cite{witten2016data}, large clusters \cite{zikopoulos2011understanding}, \cite{meng2016mllib}, \cite{mckinney2010data} or from service platforms in \CC{} \cite{hashem2015rise}.



Within the area of \DM{}, these \CC{} providers and platforms offer barely \TS{} services and methods in a catalog of services \cite{hashem2015rise}. This means having to implement specific methods and algorithms for working with \TS{} on each provider platform or to migrate all the source code developed in a specific programming language and to ensure the entire service deployment will work properly. \TS{} modeling can be a highly complex and it supposes a non-linear analysis tasks \cite{hamilton1994time} including  a set of methods of the application of \DM{} and \ML{} techniques \cite{harris2003applied},  models, evaluations of performance or precision measurements, among other, that can be seen as a workflow of tasks.


%

The services of \TS{} analysis along \CC{} service providers address a lack of integration from heterogeneous and non-standardized cloud computing platforms. When migrating services from one \CC{} provider to another, the ideal solution would be to harmonized the description of services and \WFlong{} related to \TS{} on \CC{} deployments, abstracting the programming language or the architecture of deployment. This would allow the interoperability of these services between providers to be exploited more efficiently and offer all the scalability and flexibility advantages provided by the \CC{} paradigm. With this idea we pursue the industrialization of IT services through pre-designed and pre-configured solutions that are highly automated and repeatable, scalable and reliable, by meeting the needs of users or organizations.


The aim of this paper is to propose a definition of services for \TS{} \WFlong{} in \CC{}  environments based on semantic technology, 
according to the Linked Data \cite{bizer2011linked} proposal. To address this definition, an exhaustive set of functions and algorithms related to the \TS{} and the \WFlong{} modeling have been specified and implemented (such as data pre-processing, visualization, modeling, or precision measures, among other). For the description of the different components and structures, existing service description schemes \cite{vandenbussche2017linked} have been re-used, improving the overall comprehension of the services. As a result, the \tswfschema{} proposal has been developed, a \WFlong{} definition and \TS{}  modelling, which together with \textit{dmcc-schema} \cite{parra2018data} allow to cover the complete definition of \CC{} services for \WFlong{}. 



This paper is organized as follow: in the next section (\ref{sec:relatedwork}), we present the related work within the scope of \TS{} such as its modelling and analysis, data mining \WFlong{} and aspects of \TS{} and \CC{}. Section \ref{sec:timeseries}  presents the proposed \CC{} service definition of \TS{} for \WFlong{}, including each of the elements considered for \TS{} modelling, such as seasonality analysis, prediction methods or pre-processing, among others. In the section \ref{sec:usecases}, 
several use cases are implemented. Finally, the conclusions of the work and the proposals for future work are set out in section \ref{sec:conclusions}.

\section{Related Work}
\label{sec:relatedwork}


\TSlong{} analysis and modeling is a very dynamic area that is attracting the interest of the scientific community in an incremental manner, due to the rise of IoT environments and knowledge extraction from  data sources in real time or offline. In this way, an important part of the work carried out with time series has been developed within distinct fields such as business, economics, stock market, environmental sciences, industrial monitoring, or engineering among others \cite{dominici2002use, aljawarneh2016similarity, xu2014tsaaas}.


The analysis and modeling of \TS{} is a complex task that includes the application of various operations, techniques and algorithms. This procedure can be seen as a \WFlong{}, covering a large number of methods and techniques to apply and focused on solving \TS{} problems \cite{montgomery2015introduction}.


The analysis of the time series has been studied in depth and there is no single criterion that establishes which is the procedure to be carried out for the \WFlong{} in this type of problem. There are different methodologies to tackle the problem of modelling and the approach to \TS{} resolution. The most widely used proposal is the Box-Jenkins methodology \cite{tang1991time}. This methodology can be considered as linear \WFlong{}. Box-Jenkins is used in the construction process of the ARIMA \cite{chatfield2016analysis} model for the \TS{} covering aspects such as identification, estimation, error-testing and application of methods and modelling \cite{makridakis1997arma}. Focusing  on modelling, techniques related to ARIMA, such as AR, MA or ARMA \cite{ubeyli2004spectral}, are also used as part of the \TS{} \WFlong{} process. The modelling of \TS{} from a non-linear perspective has been approached with the use of ANN \cite{zhang2003time}, bi-linear model, TAR or ARCH \cite{engle2001garch}. Other modelling techniques based on \DM{} and \ML{} have been proposed including \RFlong{} \cite{liaw2002classification}, \SVMlong{} \cite{hearst1998support}, \NNlong{} \cite{zhang2003time}, or the more modern \DLlong{} one as in \cite{yang2015deep}. The \TS{} analysis also includes a comprehensive set of \ETLlong{} (ETL) data processing tasks \cite{hoey2014system}.


\textcolor{red}{TODO: add missing [REF] and figures  in the document}
For the resolution of this type of analysis, programming languages and tools have been widely used, in addition to \DM{} platforms \cite{jovic2014overview}. These utilities include all the required components and functions for the relative \TS{} \WFlong{}.  Languages such as R (with its task-view for TS) \cite{hothorn2018cran} or Python (TS-specific libraries) \cite{vanderplas2016gatspy}, software packages such as SAS [REF] or MathLab [REF] and \DM{} environments such as KNIME \cite{berthold2009knime} or WEKA \cite{hall2009weka}, offer the tools to make effective the work with \TS{}. In all of them, it is possible to design a \WFlong{} where you can specify the application of operations, visualize the results, validate errors and apply multiple algorithms for the modelling and subsequent forecasting, classification or clustering \cite{rangra2014comparative}.


Most traditional time series analysis tools are designed to work with desktop computers and are not ready to be used in \CC{} environments. Leveraging the computing capabilities of organizations, part of that vast set of resources and infrastructure are being allocated to \DM{} and \ML{} as on-demand \CC{} services \cite{hashem2015rise}. \TS{}  analysis is no exception and more and more \CC{} providers are including specific functions and algorithms for working with \TS{} services in their catalog \cite{chen2015data}. Currently, there is a growing demand for services that allow creating   \WFlong{} to be deployed over \CC{}, such as \TS{} \cite{marozzo2016workflow,chen2014big}. These \WFlong{} are very interesting because of their scalable character existing a clear need to move much of the data processing to cloud platforms, abstracting the underlying computing infrastructure and scalability needs, both of which will be assured \cite{kranjc2015active,kranjc2012clowdflows}. 


One of the main problems of \CC{} services is the lack of a consistent and standardized definition of these services among the different \CC{} providers and it has been widely studied  \cite{dillon2010cloud, barry2003web}. This happens in the same way with the description of  \WFlong{} and experimentation with data in \CC{} in the analysis of \TS{}. The portability of services and the ability to abstract the underlying infrastructure makes it necessary to validate this type of problem on \CC{} platforms \cite{fox2009above} .

\WFlong{} modeling for \DM{} experimentation is considered in \cite{talia2013clouds}, \cite{data2013cloud} performing \WFlong{} as \CC{} services giving the user the ability to deploy a work of experimentation in an integral way. Several approaches manage the problem of the description of \WFlong{} dealing with ontology-based such us RDF or Turtle \cite{world2014rdf}. Those languages for \WFlong{} definition have been discussed in \cite{youseff2008toward} and \cite{iosup2011performance}. 

The definition and description of generic  \CC{} services that integrate multiple aspects of experimental work for \DM{} have been worked from the scope of the Linked Data \cite{bizer2011linked} proposal. Research papers such as DMOP \cite{keet2014exploring}, Exposé \cite{panov2014ontology}, dmcc \cite{parra2018data}  or MLSchema \cite{esteves2016ml} are examples of languages proposals for the definition of generic data mining services and \WFlong{}, related to \ML{}. These provide an adequate definition of services using a highly flexible definition language and linked to the natural development of the \CC{}.

Most of the proposals allow the development of experimental analysis, including part of the usual \WFlong{} with data processing and algorithms \cite{vockler2011experiences}. In these works the \WFlong{} modeling with \TS{}  is not taken into account being a fundamental element for the integration of these services within \CC{} platforms \cite{khan2012workload}. These types of problems need to bring together the different techniques and algorithms of \TS{} modeling, pre-processing \cite{garcia2015data}, performance measurements [REF], visualization [REF], or predictive models among others. In our work, a proposal of \WFlong{} modeling for \TS{} in \CC{} has been developed, which allows to tackle the work with the \TS{} experimentation and modeling using Linked Data recommendation on services description in  \CC{} platforms.

\section{Time Series service description}
\label{sec:timeseries}

Time Series modeling is a challenging task that integrates different actions following a dynamic \WFlong{}. In this work a complete proposal for \WFlong{} modeling with \TS{} in \CC{} has been made. The proposed  schema is called \tswfschema{}  and has been developed using a semantic language based on ontology, following the Linked Data guidelines for the definition and description of concepts, entities and relationships to other schemes. A complete diagram has been defined allowing any \WFlong{} with \TS{} to be modeled, as can be seen in the section on \ref{sec:usecases}, in which several examples of working with time series are developed and modelled using \tswfschema{} . 

\begin{figure}[]
  \centering
   {\epsfig{file = 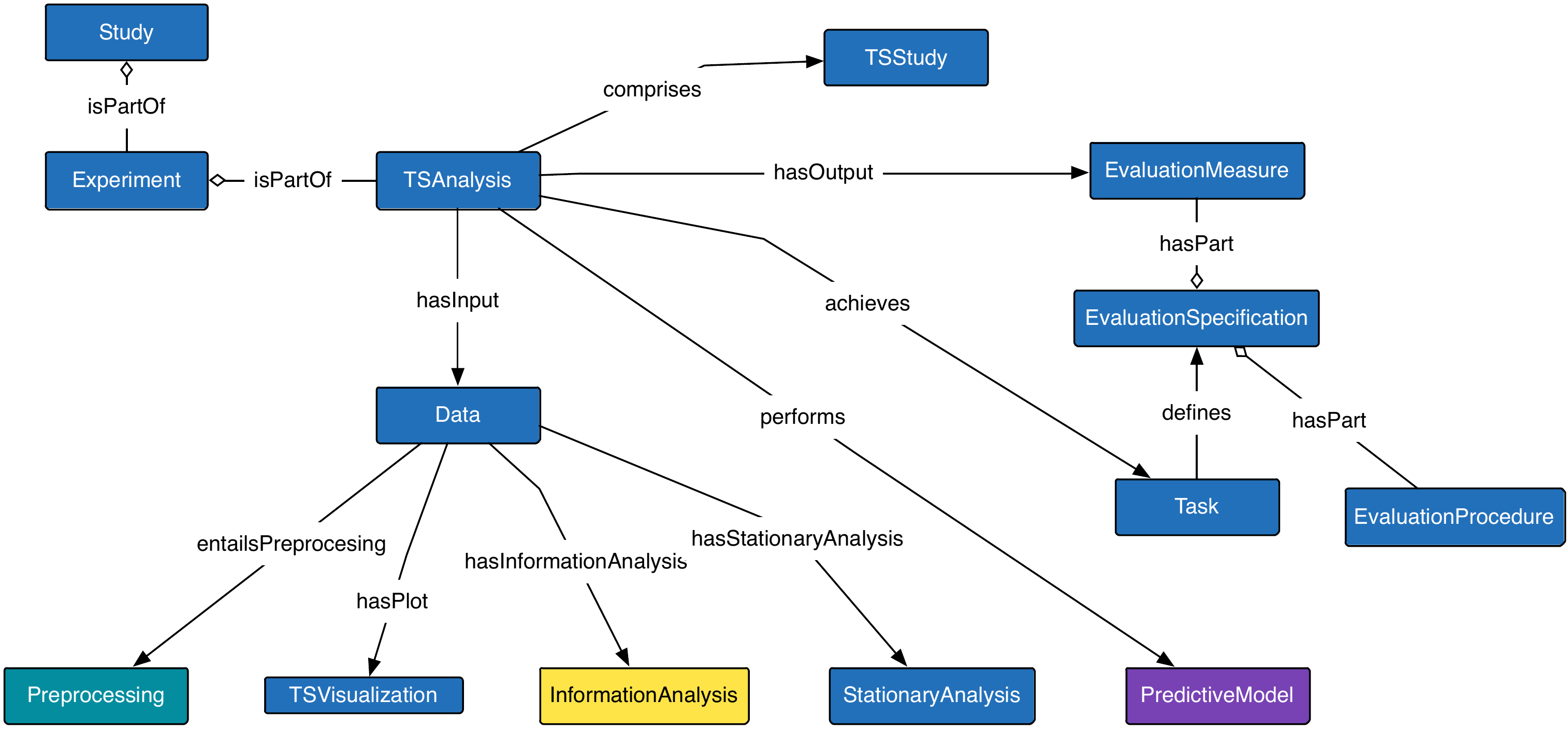, width =  \columnwidth}}
  \caption{General schema for Time Series workflow modelling}
  \label{fig:tswf-schema-general}
 \end{figure}

The scheme is divided into several parts that facilitate its integration and modularity, these are: \textit{data pre-processing}, \textit{data visualization}, \textit{functions for information analysis}, work with \textit{seasonality analysis}, \textit{predictive model selection}, \textit{learning problem} information, \textit{data entry}, and \textit{performance evaluation measures}, among others. Each of these parts is developed in detail throughout this section. In figure \ref{fig:tswf-schema-general} you can see the general diagram of the flow rate modeling with time series that has been designed. For reasons of space, the complete scheme has not been displayed, given its size, so that each of the main container classes has been shown. Following the Linked Data specification, other vocabularies have been used from other schemes, completing and broadening the definition of the scheme.  The scheme provided by \textit{MLSchema} \cite{esteves2016ml} and \dmccschema{} \cite{parra2018data} has been used as base for the \WFlong{} of experimentation with \TS{}, in addition to other vocabularies such as \textit{SKOS} \cite{miles2009skos} or \textit{schema.org} \cite{guha2016schema} among other.

For the definition of \WFlong{} modeling has been taken into account a high volume of research papers and books related to \TS{} modeling and analysis. For the container classes and other parts of the scheme, research related to the area has been reviewed, along with actual \TS{} analysis work from various sources such as \cite{harris2003applied} \cite{de1992some} \cite{hyndman2007automatic}. This has made it possible to extract a large part of the operations and methods used to model \TS{}, also from multiples knowledge domains. In the proposal developed in \tswfschema{} we have integrated the greatest number of functionalities, together with the most common ones during the \WFlong{} process with \TS{}. It also comprises widespread modeling such as the \textit{Box-Jenkings} \cite{tang1991time}  methodology as well as workflow-based experimentation for processing \DM{} problems and \ML{} applied to \TS{}  \cite{bontempi2012machine} \cite{ahmed2010empirical}.


\textit{tswf-schema} contains all key elements in the description of cloud computing services, such as interaction points, prices, instances or SLAs, among others, and serves as a complement to the integration of a \WFlong{} with time series as a Service on \CC{} platforms. This means that it allows you to have the complete description of \CC{} services, both from a functional and a business point of view.

\begin{figure}[]
  \centering
   {\epsfig{file = 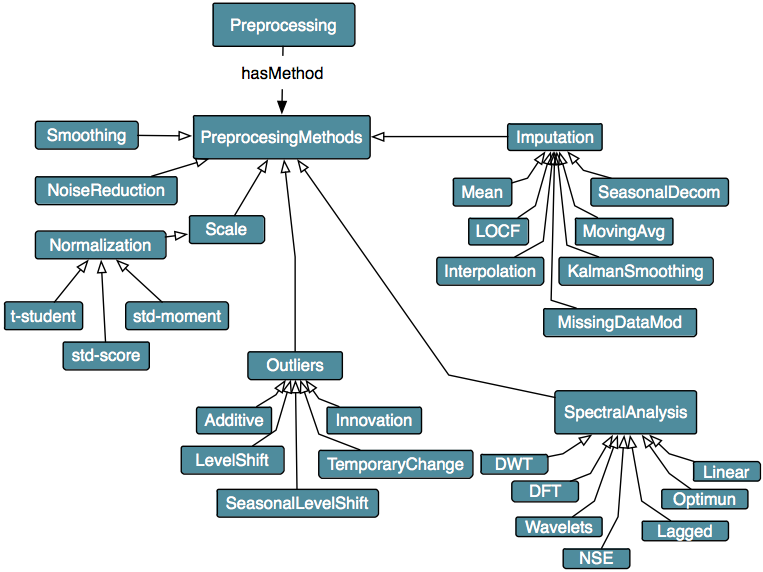, width =  \columnwidth}}
  \caption{Preprocessing classes for Time Series workflow modelling}
  \label{fig:tswf-schema-preprocesing}
 \end{figure}

In the following subsections, all the main components of the \textit{tswf-schema} definition are detailed:


\textbf{Pre-processing.} Part of the analysis and work with \TS{} requires operations on the data, where they apply transformations, reductions, cleaning, or imputations among others. In the diagram in figure \ref{fig:tswf-schema-preprocesing} you can see all the elements most commonly used in data level time series processing. It has basically been segmented into various parts, such as imputation, \textit{outliers}, spectral analysis, scaling, noise reduction or smoothing. Each of the sub-parts contains several of the functionalities that have been considered the most commonly used in the processing of these types of data, according to the work of \cite{garcia2015data}.

\begin{figure}[]
  \centering
   {\epsfig{file = 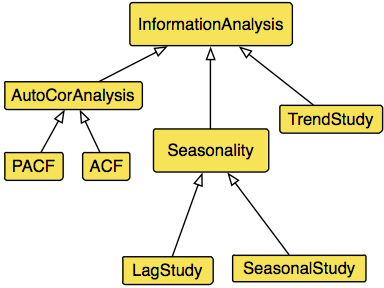, width = 5cm}}
  \caption{Information analysis classes for Time Series workflow modelling}
  \label{fig:tswf-schema-informationanalysis}
 \end{figure}

\begin{figure}[]
  \centering
   {\epsfig{file = 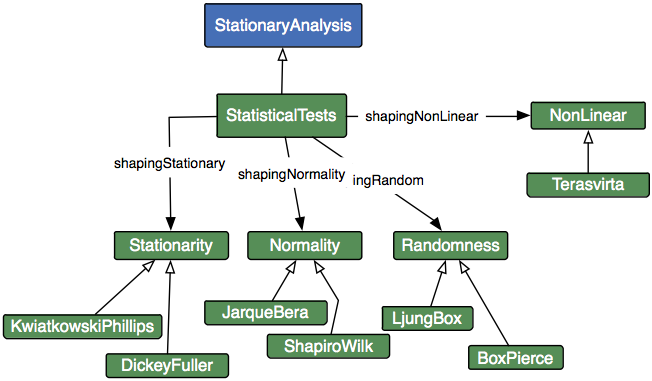, width =  \columnwidth}}
  \caption{Stationary analysis classes for Time Series work-flow modelling}
  \label{fig:tswf-schema-stationalanalysis}
 \end{figure}

\textbf{Analysis of the information and analysis of seasonality.} Most of the statistical studies with \TS{} are based on the Box-Jenkins methodology \cite{tang1991time}, where a series of studies on the data of the time series is applied to check the trend, seasonality among others, as well as different statistical tests necessary to correctly identify the time series. As shown in figures \ref{fig:tswf-schema-informationanalysis} and \ref{fig:tswf-schema-stationalanalysis}, they have been divided into two parts, on the one hand, the analysis of the information, which includes correlation, seasonality and trend tests. On the other hand, it also includes statistical tests, integrating stationarity, normality, randomness or non-linearity among others.

\begin{figure}[]
  \centering
   {\epsfig{file = 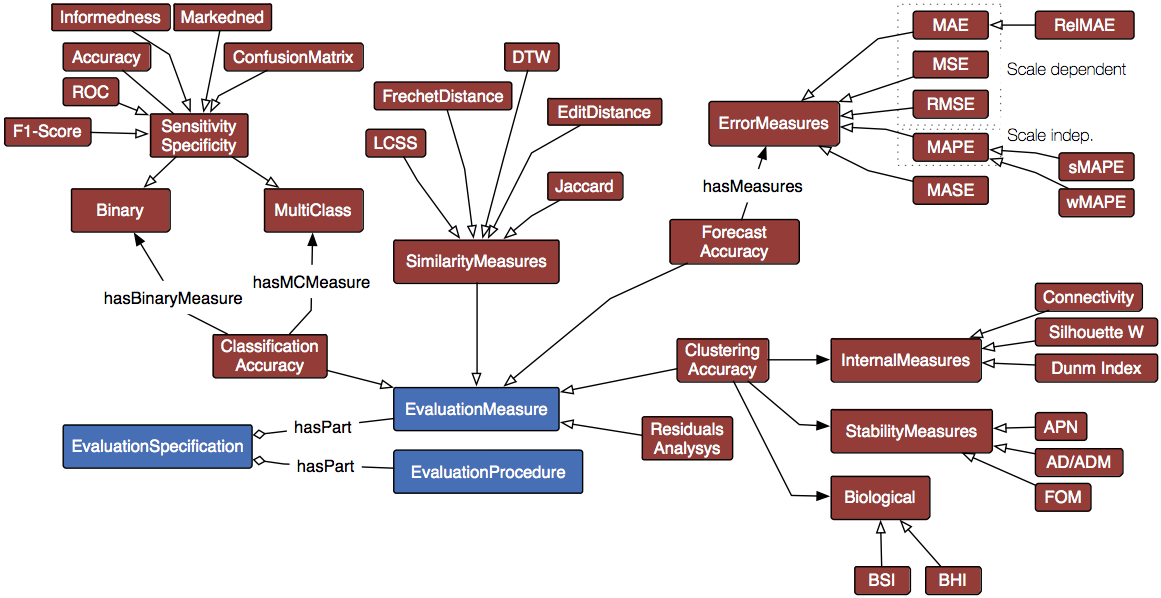, width = \columnwidth}}
  \caption{Evaluation measures classes for \TS{} workflow modelling}
  \label{fig:tswf-schema-evaluationmeasures}
 \end{figure}

\textbf{Evaluation measures.} The output of the analysis \WFlong{} of a \TS{} can be identified according to the \TS{} study problem in question. Four main groups of measures have been considered, related to those most used in \TS{} problems, such as classification, similarity measurements \cite{serra2014empirical}, precision of the forecast \cite{weigend2018time}, performance of the clustering \cite{montero2014tsclust}, or analysis of the residues. Classification measures such as \textit{F1-Score}, \textit{ROC}, or matrix confusion have been taken into account. For similarity measures, others such as \textit{DTW} \cite{mueen2016extracting}, \textit{Edit Distance} \cite{chen2015kernel} or \textit{Jaccard} \cite{aghabozorgi2015time} have been implemented, as they are common in problems of this type. As regards the \textit{Error measures}, which identify the quality of the forecast adjustment, all the measures considered in the work \cite{shcherbakov2013survey} have been included. If the problem being addressed is related to time series clustering, different measures such as \textit{APN}, \textit{AD/ADM} or \textit{Silhouette W}  have been taken into account. Figure \ref{fig:tswf-schema-evaluationmeasures} details the set of evaluation measures that have been included in \tswfschema{}.

\textbf{Predictive model.} When instantiating a model that tries to fit the data of the \TS{}, a multitude of algorithms and statistical methods can be applied. In our work we have integrated the algorithms and methods most commonly used in this area of \TS{} data analysis. From widespread statistical methods such as \textit{ARIMA} or variants of it (\textit{ARIMA}, \textit{SARIMA}, \textit{ARIMAX}, etc.), as well as others such as \textit{ETS} among others. Regression analysis has also been implemented with methods related to time series considered in \cite{paolella2018linear}, such as \textit{AR}, \textit{LASSO} or \textit{MARS} among others. Finally, an important part of Machine Learning methods has been considered, mainly from \cite{de200625} studies. These algorithms include the application of techniques based on \textit{Neural Networks}, \textit{Random Forest} or \textit{SVM} among the most outstanding. The complete diagram of the methods and algorithms included in \tswfschema{} can be seen in figure \ref{fig:tswf-schema-evaluationmeasures}.

 \begin{figure}[]
  \centering
   {\epsfig{file = 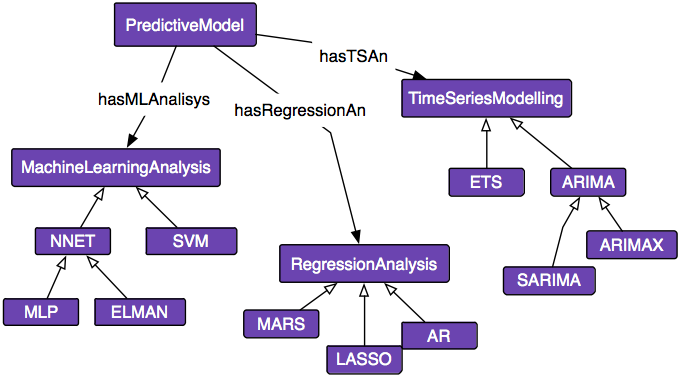, width =  \columnwidth}}
  \caption{Predictive models classes for Time Series work-flow modelling}
  \label{fig:tswf-schema-predictivemodel}
 \end{figure}

\begin{figure}[]
  \centering
   {\epsfig{file = 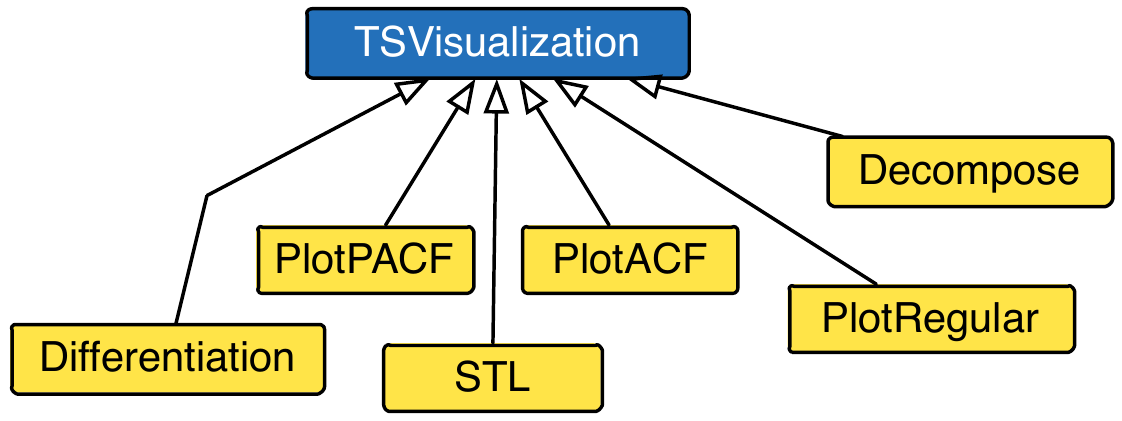, width =  5cm}}
  \caption{Visualization classes for Time Series work-flow modelling}
  \label{fig:tswf-schema-visualization}
 \end{figure}

\textbf{Visualization of the data.} One part of the work is to perform repeated visualizations of the data to assess the state of the data and to use different plot analyses to better understand the shape of the data. This allows to observe some of the visual information such as decomposition, differentiation or \textit{STL} plots. Figure \ref{fig:tswf-schema-visualization} implements the set of visualizations available.

\textbf{Workflow.} The study of data generally does not follow a linear work scheme as indicated with this particular type of data, so that the data engineer or data scientist generates different flows of operations with which to compose the work during the analysis. To facilitate this task in \tswfschema{}, all operations and tasks that are performed can be composed as a linear or nonlinear sequence. If we think at a high level, this feature is what will provide the \CC{} service for \TS{} with the necessary functionality to visually compose (as building blocks) each operation or task to be performed at each time to produce a complete analysis. In figure X can be seen how this functionality intrinsic to \tswfschema{} serves to manage the flow of operations contained in the scheme.


\section{Deploying Time Series services}
\label{sec:deploying}

Once all the components of \tswfschema{} have been defined in the previous section, it is necessary to present a set of examples that highlight the potential of the scheme designed to describe \TS{} workflows. 

To address the presentation of these cases, two models will be used, on the one hand a) the straightforward transcription of examples proposed in a programming language into \tswfschema{}, putting the value on the ease of translating any \TS{}-based study into a semantic agnostic model ready to be ported, deployed or executed on \CC{} providers, and b) the modelling of a cloud service with key components of the service, so that a \CC{}  service is specified and it is ready to be part of a \CC{} provider's catalogue. For the examples we will use \R{} and \python{} language, since they are one of the most extended languages and for the transcription of those example codes to the \tswfschema{} scheme we will use \jsonld{}.

\begin{figure}[]
  \centering
   {\epsfig{file = 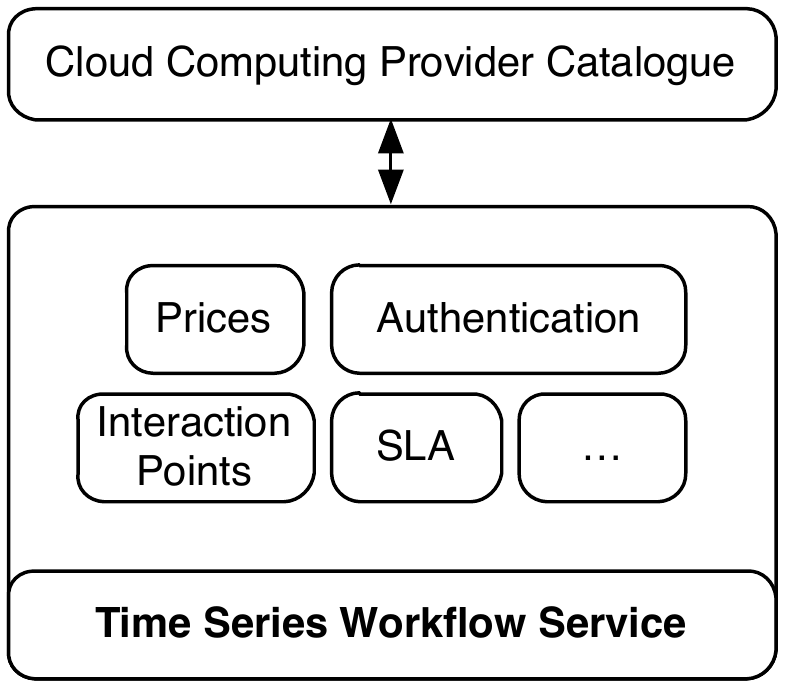, width =  0.5\columnwidth}}
  \caption{Components of a generic \CC{} service together with the functionality elements of a \TS{} service in a service catalogue.}
  \label{fig:cloudcomputinservice}
 \end{figure}

In order to make the process of developing a \CC{} service based on \TS{} more comprehensible, the figure \ref{fig:cloudcomputinservice} shows the different elements that make up a \CC{} service. This figure  shows several aspects related to the definition of the service management logic, more closely linked to the definition of \CC{}, and on the other hand the description of the service functionality itself, in this case, a workflow with time series that runs as a service within a \CC{} provider's service catalogue.

\subsection{Baseline of the structure of a \TS{} workflow}

The general minimum structure for designing and instantiating a \TS{} analysis using \tswfschema{} can be seen in the listing \ref{lst:lakehuron_skeleton}, according to the figure \ref{fig:tswf-schema-general} that outlines the overall diagram of the scheme. This skeleton contains the basic components that will be instantiated to show a functional example of the description of a workflow with \TS{}, where in line 2-3 is established which will be the context of vocabulary used for the definition in \jsonld{}, from line 6 to 10 are added the different operations of pre-processing of input data and various studies on them. Then in lines 11 and 12 are stated the methods that will be used to create the models and from line 15 the workflow information output. Each of the skeleton components is detailed in the following sections.

\footnotesize{
\begin{lstlisting}[caption={Skeleton of the component instantiation in \jsonld{}},label={lst:lakehuron_skeleton},language=json,firstnumber=1]
{
 "@context": {
    "tswf": "http://dicits.ugr.es/linkeddata/tswf-schema/"
  },
  ...
  "tswf:hasInput": {
    "tswf:hasPlot": {...},
    "tswf:hasInformationAnalysis": {...},
    "tswf:hasStationaryAnalysis": {...}
   },
  "tswf:performs": {
    "tswf:hasTSAnalysis": {...},
    ...
    }
  "tswf:hasOutput": {
  
  },
  ...
 }
\end{lstlisting}
}

\subsection{A model of \TS{} analysis}

\begin{figure}[]
  \centering
   {\epsfig{file = 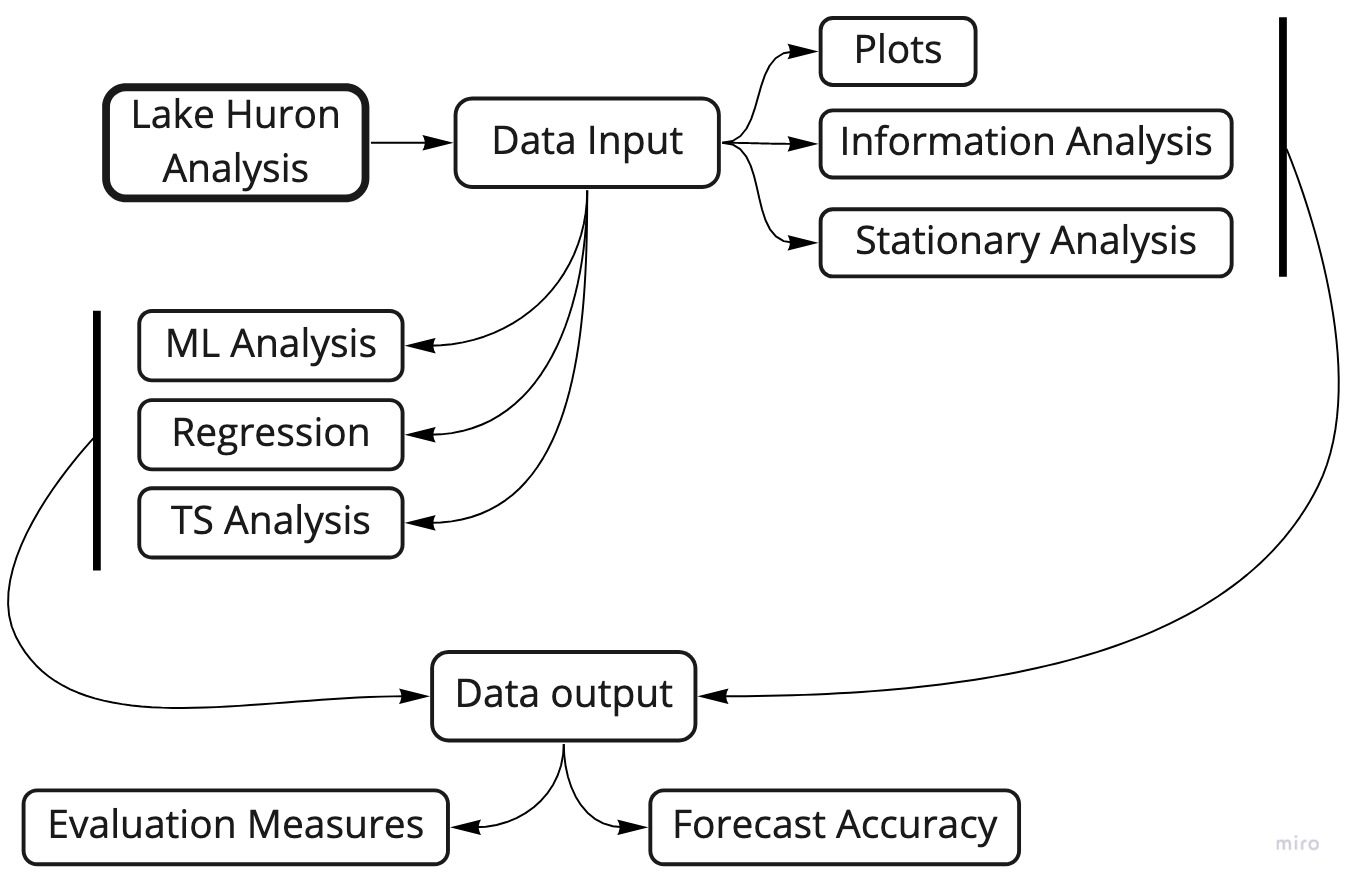, width =  \columnwidth}}
  \caption{Diagram of the operations and tasks carried out for the case study use of Lake Huron.}
  \label{fig:lakehuronhighleveloperative}
 \end{figure}

For this use case we will use an example of an actual time series analysis, implemented in the \R{} language. This example consists of a series of basic steps that reproduce the modeling of a simple \TS{} example. This workflow can be seen visually in the figure \ref{fig:lakehuronhighleveloperative}.

The full code in \R{} and \python{} of the workflow implementation can be found in the scheme repository \cite{RepoDicitsTSWF}. In the description of the use case, we represent some parts of the modeling as a service, so we can compare the \R{}-language code transcription to the \tswfschema{} scheme with \jsonld{} as a semantic description language. This is not to replace one language with another, but with \tswfschema{} a complete description of a high level workflow is provided, abstracting the programming support, the platform or the computing environment. One of the advantages of using semantic models is that you can indicate the level of detail you want in each defined aspect. This is more expressive and less complex to understand, leaving more core aspects to the implementation of the underlying system that manages the schema, such as the default settings.

\begin{figure}[]
  \centering
   {\epsfig{file = 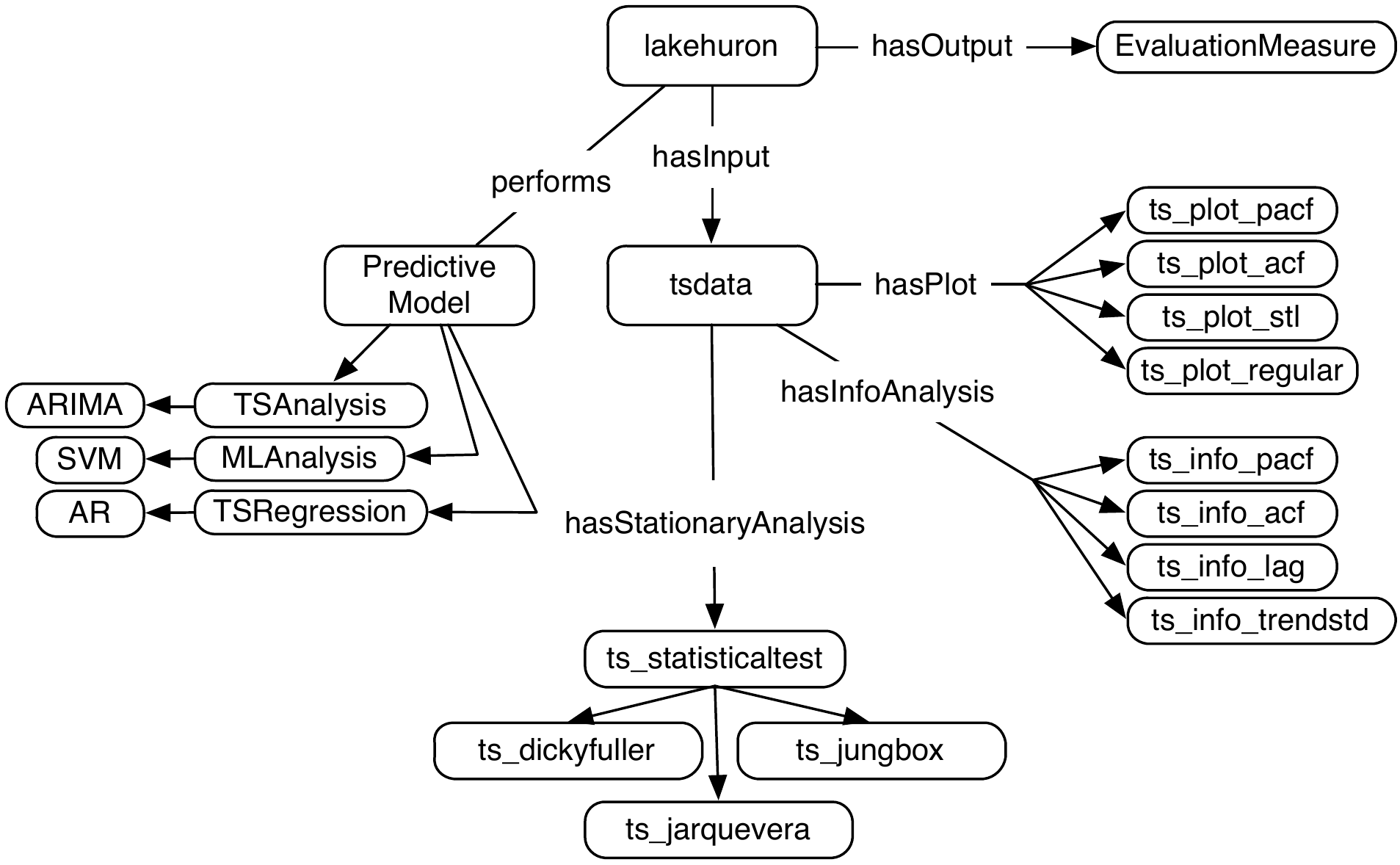, width = \columnwidth}}
  \caption{Diagram of the components and operations of a simple TS analysis use case using \tswfschema{}.}
  \label{fig:lakehurontswf}
 \end{figure}
 
 One of the advantages of using semantic description technologies to define services or functions is that you can indicate the level of detail you want in each defined aspect. This is more expressive and less complex to understand, leaving more core aspects to the implementation of the underlying system that manages the schema, such as the default settings.

\subsubsection{\TS{} Analysis: Workflow information and description}

The definition of services in general terms requires a minimum of basic information describing what is being implemented. In our case, the first step to define the workflow instantiation with \tswfschema{} is the description of the analysis, through the main class \texttt{TSAnalysis}.With \texttt{TSAnalysis} and its properties all the attributes considered key in the definition of what is going to be deployed are declared so that it can be described in the most efficient way possible and with a friction-less integration in catalogs or marketplace of services in Cloud Computing. Aspects related to the automatic discovery of services for \CC{} would be defined from this class. For space reasons not all attributes of \texttt{TSAnalysis} will be exposed, so only some basic ones are shown in the listing \ref{lst:lakehuron_TSAnalysis}.

\footnotesize{
\begin{lstlisting}[caption={Definition of the basic information of time series analysis.},label={lst:lakehuron_TSAnalysis},language=json,firstnumber=1]
"@context": {
    "tswf": "http://dicits.ugr.es/linkeddata/tswf-schema/"
  },
  "@id": "http://dicits.ugr.es/tswf-marketplace/#TS_eb09t74",
  "@type": "tswf:TSAnalysis",
  "tswf:name": "TS Analysis base with code TS_eb09t74",
  "tswf:description": "This is an example of TS an.",
  "tswf:author": "This is an example of TS an.",
  "tswf:dateCreated": "2020-09-01 10:30:00",
  "tswf:version": "Test version 1.0",
  "tswf:codeRepository": {
        "@type":"tswf:url",
        "@value":"https://"
        } 
  "tswf:hasInput": ...
\end{lstlisting}
}

In the first lines (lines 1 to 3) the context of the overall \texttt{TSAnalysis} description is declared in which \tswfschema{} will be used. The context allows applications to use a set of terms to communicate with one another more efficiently, but without losing accuracy. Then, each instance of \texttt{TSAnalysis} created with \tswfschema{} must be marked with an identifier that uniquely relates this workflow, for example, within a particular catalog, for which \texttt{@id} is specified (line 4). The \texttt{@type} property indicates that all the remaining content corresponds to a \texttt{TSAnalisys} and subsequently part of its attributes are declared (lines from 5 to 13), such as the \TS{} name, description of the workflow, or the repository where the code of the described service will be hosted, to take the description of the flow to execute it (lines 11-14). The rest of the properties of the workflow components are instantiated in the following subsections (from line 15, with "\texttt{tswf:hasInput}", "\texttt{tswf:hasOutput}", etc. as described within  figure X).

\subsubsection{Data input}

To perform the ingestion of the data of a time series, basically in the \R{} language the data can be loaded from different sources, such as \textit{CSV} format for example. In \CC{} environments or other computing platforms it is possible that data can come from databases, \TS{} databases, or even data steaming services.  In this way,  just as \R{} or \python{} supports data ingest from multiple sources, \tswfschema{} has the possibility to include very diverse data sources. For the data input instantiating it is necessary to at least define the data source, the source type and the fields that will be used during the whole analysis process. In the listing \ref{lst:lakehuron_hasInput}, a simple instantiation of the data entry is shown, where over the line number \texttt{2} is indicated the type of entry (\texttt{tswf:Data}), the type of source (\texttt{tswf:CSVFile}  in this case) in lines \texttt{3-4} and where physically are the data by means of the property \texttt{tswf:src} (line 5). Then, with \texttt{tswf:fields} the features and fields of the data source are set within the lines \texttt{6-12}.

\footnotesize{
\begin{lstlisting}[caption={Instantiation of the data entry source and its parameterization.},label={lst:lakehuron_hasInput},language=json,firstnumber=1]
...
"tswf:hasInput": {
  "@type": "tswf:Data",
  "tswf:source": {
    "@type": "tswf:CSVFile",
    "tswf:src": "///dicits/examples/lakehuron.csv",
    "tswf:fields": {
       "@set": [
         {"@value": "Year",
          "@type": "tswf:datetime"},
         {"@value": "Level",
          "@type": "tswf:integer"}
        ]
      }
    },
    ...
\end{lstlisting}
}
The transformation of the initial data intake part into \tswfschema{} is done at a higher level, since semantically each operation is labeled with its properties. This allows not having to define in advance each detail of the parameterization (or indeed the application of the \TS{} type), so that this complexity can be left to the underlying system or platform as the service responsible for making these tweaks. 

\subsubsection{Preliminary exploratory analysis}

A basic part of the analysis of \TS{} data is exploratory data analysis where the data can be plotted, checked for patterns or trends, verified for seasonality or the presence of cycles, among others. In the example, we are dealing with, through the input data we make several key charts to visually understand the problem. We consider for instance \pacf{} (\textit{Partial Autocorrelation Function}), \acf{} (\textit{Autocorrelation Function}), \stl{} (Seasonal and Trend decomposition using \textit{Loess}) as well as a visual representation of the data. 

The graphical representation of the input data is done from the \texttt{tswf:hasPlot} property. In \tswfschema{} there is a set of possible definitions of the most commonly used diagrams and charts within \TS{} analysis, so to instantiate a type of chart that you want to display is only needed to include the list of types that are required for the analysis as shown in listing \ref{lst:lakehuron_hasPlot}. Note that the level of detail selected to instantiate these types of chars does not include parameterization except in \texttt{tswf:PlotPACF}, this is because the internal implementation will be in charge of using the correct visualization with the default parameters of each chart type for the input data. On the other hand, for \texttt{tswf:PlotPACF} a \texttt{lag} parameter (\texttt{tswf:parameters}) and its specific value have been indicated (lines from 6 to 12).

\footnotesize{
\begin{lstlisting}[caption={Set of diagrams/charts that are deployed},label={lst:lakehuron_hasPlot},language=json,firstnumber=1]
...
"tswf:hasPlot":{
      "@type": "tswf:TSPlot",
      "@set":[
        {"@type":"tswf:PlotSTL"},
        {"@type":"tswf:PlotACF"},
        {"@type": "tswf:PlotPACF",
          "tswf:parameters": {
            "@set": [
              {"tswf:name":"lag", "@value": 10}
            ]
          }
        },
        {"@type":"tswf:PlotRegular"}
      ]
    },
    ...
\end{lstlisting}
}

\footnotesize{
\begin{lstlisting}[caption={\TS{} information analysis.},label={lst:lakehuron_hasInformationAnalysis},language=json,firstnumber=1]
...
"tswf:hasInformationAnalysis": {
      "@type": "tswf:InformationAnalysis",
      "@set": [
        {"@type": "tswf:LagStudy"},
        {"@type": "tswf:TrendSTL"},
        {"@type": "tswf:ACF"},
        {"@type": "tswf:PACF"}
      ]
    },
    ...
\end{lstlisting}
}

Further on, in the listings \ref{lst:lakehuron_hasInformationAnalysis} and \ref{lst:lakehuron_hasStationaryAnalysis} are shown how it is possible to add  different studies used in the analysis, such as \textit{Information Analysis}, or \textit{Stationary Analysis}. For both types of studies, following the scheme shown in figures \ref{fig:tswf-schema-informationanalysis} and \ref{fig:tswf-schema-stationalanalysis}, it is possible to include their corresponding tests and operations. For the analysis of the information, as shown in list \ref{lst:lakehuron_hasInformationAnalysis}, four basic studies have been added in the work with \TS{} (lines 4 to 8).  In the same way as with the descriptions of the charts, we have taken all the default parameters for this example. On the other hand, in the study of stationarity analysis shown in list \ref{lst:lakehuron_hasStationaryAnalysis} four basic tests have been added (lines 4 to 8).

\footnotesize{
\begin{lstlisting}[caption={\TS{} stationary analysis.},label={lst:lakehuron_hasStationaryAnalysis},language=json,firstnumber=1]
...
"tswf:hasStationaryAnalysis": {
      "@type": "tswf:StatitionaryAnalysis",
      "@set": [
        {"@type": "tswf:StatisticalTest"},
        {"@type": "tswf:DickeyFuller"},
        {"@type": "tswf:JarqueBera"},
        {"@type": "tswf:JungBox"}
      ]
    },
    ...
\end{lstlisting}
}

\subsubsection{Predictive model and evaluation measures}

Once the exploratory analysis is done, the selection and adjustment of the model is performed. The selection of the best model depends on the availability of historical data or the dependence of variables. Hence it is often necessary to compare several predictive models to see how the data actually behave. For this example the \R{} code of the three models displayed in listing \ref{fig:lakehuronhighleveloperative}, where an \texttt{AR} regression model, an \ARIMA{} modeling and finally an \SVM{} algorithm are used. The consequent code in the \tswfschema{} scheme corresponds to the listing  code \ref{fig:lakehuronhighleveloperative}. For each of the methods,  it is possible to specify the parameterization, in this case only for the \ARIMA{} model has been specified. In listing \ref{lst:lakehuron_performs}, from line 2 to line 5, a model based on Machine Learning functions/operations is specified, such as \texttt{SVM} (\texttt{tswf:SVM}). Another regression-based model is indicated in lines 10-12 (\texttt{tswf:AR}) and finally, with \texttt{tswf:ARIMA} (lines 7-15) an \texttt{ARIMA} model is instantiated, with a partial parameterization (\texttt{tswf:parameters}) as it appears in lines 11-13 including \texttt{order}, \texttt{seasonal} and \texttt{lambda}.

\footnotesize{
\begin{lstlisting}[caption={\TS{} predictive model.},label={lst:lakehuron_performs},language=json,firstnumber=1]
...
"tswf:performs": {
      "@type": "tswf:PredictiveModel",
      "tswf:hasMLAnalysis": {
        "@type": "tswf:SVM"
      },
      "tswf:hasTSAnalysis": {
        "@type": "tswf:ARIMA",
        "tswf:parameters": {
          "@set": [
            {"tswf:name": "order", "@value": [0,0,1])},
            {"tswf:name": "seasonal", "@value": [0,0,1])},
            {"tswf:name": "lambda", "@value": 0)}
          ]
        }
      },
      "tswf:hasTSRegression": {
        "@type": "tswf:AR"
      }
    },
...
\end{lstlisting}
}

Finally, as part of the explicit output of the workflow, using the \texttt{tswf:hasOutput} property, it is possible to indicate what kind of additional information can be analyzed as a result of the full process (listing \ref{ls:lakehuron_evaluationmeasures}, line 2). This has included in the output of the workflow the prediction function, which is called to return the results on how is the forecasting on different horizons, by giving the user of different measures to evaluate the performance of the forecasts, for example (see figure \ref{fig:tswf-schema-evaluationmeasures}). In the listing  \ref{ls:lakehuron_evaluationmeasures}, lines 5-10 we have chosen to include two measures (RMSE and MSE) for the evaluation (\texttt{tswf:EvaluationMeasures}) with which you can verify the accuracy of the forecast (\texttt{tswf:ForecastAccuracy}), made with the models generated in the previous stage through the property \texttt{tswf:performs} (in this instance \texttt{AR}, \texttt{ARIMA} and \texttt{SVN}).

\footnotesize{
\begin{lstlisting}[caption={\TS{} evaluation measures.},label={lst:lakehuron_evaluationmeasures},language=json,firstnumber=1]
...
 "tswf:hasOutput": {
      "@type": "tswf:EvaluationMeasures",
      "@set": [
        {
          "id": "tswf:TSFCastAccu",
          "@type": "tswf:ForecastAccuracy",
          "tswf:hasMeasures": [
            {"@type": "tswf:RMSE"},
            {"@type": "tswf:MSE"}
          ]
        }
      ]
    }
...
\end{lstlisting}
}

\subsection{A \CC{} service for \TS{}}
\label{subsec:ccserviceforts}

The \TS{} example provided in the section \ref{sec:timeseries} is defining a workflow as if we were doing it in a programming language like R or Python, but using \jsonld{} as a description language. With \tswfschema{} it is possible to package \TS{} workflows and make them portable between different computing platforms by reading the description and operating each high-level component on an underlying platform or by using a \textit{Broker} to decide which \textit{SVM} or \textit{ARIMA} implementation at the low-level will be used to process a particular element. This gives an idea of the potential of using this type of semantic technology to tackle the deployment of services in \CC{}. In this manner for this use case we want to put the value how \texttt{tswf-schema} can be integrated to support a \CC{} service for \TS{}. 

Following the principles of \textit{Linked Data} for semantic data, we will reuse another scheme called \texttt{dmcc-schema} \cite{parra2020semantics} that allows unifying all the basic aspects of the definition and management of a \CC{} service including costs/prices, catalog or \textit{SLA} among others, together with the functionality such as a \TS{} analysis.

   \begin{figure}[ht]
    \centering
    \begin{pgfpicture}
        \pgftext{\pgfimage[width=\columnwidth,height=5cm]{dmcc+tswfcc.png}}
        \label{fig:DMCCTSWF}
    \end{pgfpicture}
    \end{figure}

In figure \ref{fig:DMCCTSWF} a diagram can be seen combining both schemes to complete the description of a \TS{} workflow with the management of a \CC{} service.

\section{Validation}



In addition to giving in the previous section of examples of \TS{} modeling using \tswfschema{}, to reinforce the validity of the scheme as a mechanism for the definition of \TS{} analysis services in \CC{}, we have carried out two additional validation actions, such as \marketplace{} of \TS{} services for \CC{} with \tswfschema{} and a set of competence questions (\CQs{}) .

\subsection{\TS{} \marketplace{}}  


Dealing with semantic definitions and descriptions, a test of the validity of the technology for practical purposes is to deploy these components effectively within service catalogs. These catalogs have a twofold function, on the one hand, to serve as a repository of service definitions and on the other hand to provide a point of service discovery by programmatic entities having the ability to understand the services exposed. An example of a service catalog with descriptions that can be imported, exchanged, discovered, and consumed has been designed as part of a \TS{} \marketplace{}. The design of this  \marketplace{} contains 3 key features:

\begin{itemize}
    \item \textbf{Import}, where the platform allows users to include their own \tswfschema{}
    instances in \jsonld{} or \textit{\ttl{}} format. With this tool, service descriptions are
    validated and included in the catalog. Once in the catalog, it is possible to use the composition tool where the workflow description can be modeled or viewed using the \tswfschema{} elements in a visual way. Figure \ref{fig:importflow} shows the workflow import diagram to the \TS{} \marketplace{} and a part of the service catalog of available \TS{} containing all the basic attributes and details deployed. For each \TS{} it is possible to explore the definition and the components that integrate it, as well as the possibility to share/download each service available in the catalog.
    
    \begin{figure}[ht]
    \centering
    \begin{pgfpicture}
        \pgftext{\pgfimage[width=\columnwidth,height=5cm]{importflow.png}}
        \label{fig:importflow}
    \end{pgfpicture}
    \end{figure}
    
   \item \textbf{Composition},  similar to other platforms, composition offers a complementary visual tool for creating/modifying \TS{} workflows and services. In this sense it makes it much easier to compose workflows in general, since visually it is much more comfortable to use a visual tool than to code it manually. An example of the platform showing a workflow with example \TS{} is shown in the Figure \ref{fig:composeflow}. In this figure a set of  workflow operations that the service is capable of deploying are depicted. 
   
   \begin{figure}[ht]
    \centering
    \begin{pgfpicture}
        \pgftext{\pgfimage[width=\columnwidth,height=5cm]{composeflow.png}}
        \label{fig:composeflow}
    \end{pgfpicture}
    \end{figure}

    \item  \textbf{Service consumption}, once the descriptions of \tswfschema{} instances are published, they are available for consumption either manually or automatically, for example, to deploy them from \CC{} providers.
\end{itemize}

The platform is available for use in production in the website of the project \cite{noauthor_datamining-ld_nodate} and the source code is available in a \textit{Github}  repository \cite{noauthor_dicits/tswf-marketplace_2020}, in this way it is possible to deploy the \TS{} \marketplace{} in any other provider of \CC{} services.


\subsection{Competency Questions (CQs)} 
Competency questions (CQs)~\cite{uschold1996} are used to specify the knowledge that has to be entailed in the ontology/vocabulary and thus can be considered to be requirements on the content of the ontology.
A way to validate a semantic schema is the creation of a series of \CQs{} to test whether the \tswfschema{} correctly fits the problem domain and is able to accurately solve the queries made to it. With these \CQs{} we try to cover an important part of the definition of a \TS{} service in \CC{} , considering key elements such as, the study, analysis, or predictive models and also the management of the workflow of the \TS{} itself within the \CC{} provider (aspects like \CC{} service management is shown in section \ref{subsec:ccserviceforts}). The selected 10 \CQs{} are the following: 

\begin{description}
\item [CQ01] \textit{How many operations [tasks] does the X \TS{} workflow for predictive analysis have?}\\Response: .

\item [CQ02] \textit{What services from the catalog [of a \CC{} provider] enable the use of \TS{} processing functions?}\\Response: .

\item [CQ03] \textit{Does the \TS{} service provide algorithms [functions] for predictive analysis using Deep Neural Networks and does it provide an SVM algorithm?}\\Response: .

\item [CQ04] \textit{What data input does the \TS{} workflow need with the X-identifier?}\\Response: .

\item [CQ05] \textit{What are the outputs of the X-identifier of a \TS{} workflow for the \CC{} service?}\\Response: .

\item [CQ06] \textit{What is the estimated economic cost of running a \TS{} workflow analysis for the \CC{}  provider X?
}\\Response: .

\item [CQ07] \textit{Is there authentication for the execution of the \TS{} service for provider X?}\\Response: .

\item [CQ08] \textit{What is the parameterization of the ARIMA algorithm?}\\Response: .

\item [CQ09] \textit{Can you display the prediction data for a horizon of X days for the Y workflow?}\\Response: .

\item [CQ10] \textit{Which predictive model of the analysis produces the results with the lowest RMSE error?}\\Response: .

\end{description}

The aim is to confirm that with \tswfschema{} is possible to capture  partially the  features  of a specific \TS{} service and it can be integrated into a service definition, through which make queries and manage this type of services for multiple \CC{} providers. Dataset, \CQs{} and \textit{SparQL} queries (endpoint) are available on the project website \cite{noauthor_datamining-ld_nodate}.

 \section{Conclusions and future work}
 \label{sec:conclusions}

 In this article we have introduced a scheme for the description of services in \CC{} specifically designed for \TS{} processing named \tswfschema{}.  The scheme developed brings together all the commonly used operations in the study and analysis of \TS{}, which allows transforming any implementation already developed in languages such as \textit{R} or \textit{Python} into  a description based on semantic technology much richer, more homogeneous, and portable. This portability is key in terms of the fact that a single definition of a workflow with \tswfschema{} would be manageable and deployable in any \CC{} service provider significantly facilitating the industrialization of services for \CC{} environments.
 
 With the scheme developed together with other complementary ones such as \texttt{dmcc-schema}, it is possible to include in the definition \TS{} workflow all the elements related to the management of a \CC{} service such as prices, SLA or instances, among other, taking advantage of the \textit{Linked Data} proposal.  
 
 The selected use cases allow, on one hand, to show the user that any workflow implementation can be transformed into a semantic specification with \tswfschema{} ready to be consumed in \CC{} environments, and on the other hand, it highlights the need to have a homogenization in the definition of this type of computing services, which unfortunately the \CC{} service providers market lacks. In addition, thanks to CQs it is possible to respond to a subset of desirable features that a service engineer in Cloud Computing might need. Both effectiveness and efficiency have been highlighted in the uses case validation.  
 
 Finally, as a future work, we propose the implementation of a Cloud Computing services broker for \TS{} that has the ability to optimize a workflow written in \tswfschema{} taking advantage of algorithm implementations and computing capabilities over different \CC{} platforms.


%

\ifCLASSOPTIONcompsoc
  \section*{Acknowledgments}
\else
  \section*{Acknowledgment}
\fi

M. Parra-Royon holds a "Excelencia" scholarship from the Regional Government of Andaluc\'ia (Spain). This work was supported by the Research Projects \textit{P12-TIC-2958}  and \textit{TIN2016-81113-R} (Ministry of Economy, Industry and Competitiveness - Government of Spain).

\ifCLASSOPTIONcaptionsoff
  \newpage
\fi



%

\vfill

\bibliographystyle{IEEEtran}
\bibliography{IEEEabrv,bibliography/references}


%





\begin{IEEEbiography}[{\includegraphics[width=1in,height=1.25in,clip,keepaspectratio]{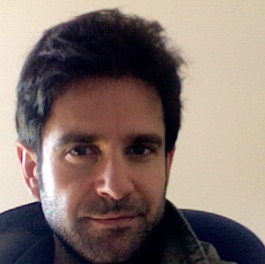}}]{Manuel Parra-Royon}
is  Computer  Engineer (2012) from the University of Granada (Spain). He received the Master’s Degree in Information Science and Computer Engineering in 2015 from the University of Granada, Spain. He is currently a Ph.D. candidate in the Department of Computer Science and Artificial Intelligence, University of Granada. His research interests include cloud computing services, data mining, machine leaning,  time series and  large-data processing on distributed computing.
\end{IEEEbiography}

\begin{IEEEbiography}[{\includegraphics[width=1in,height=1.25in,clip,keepaspectratio]{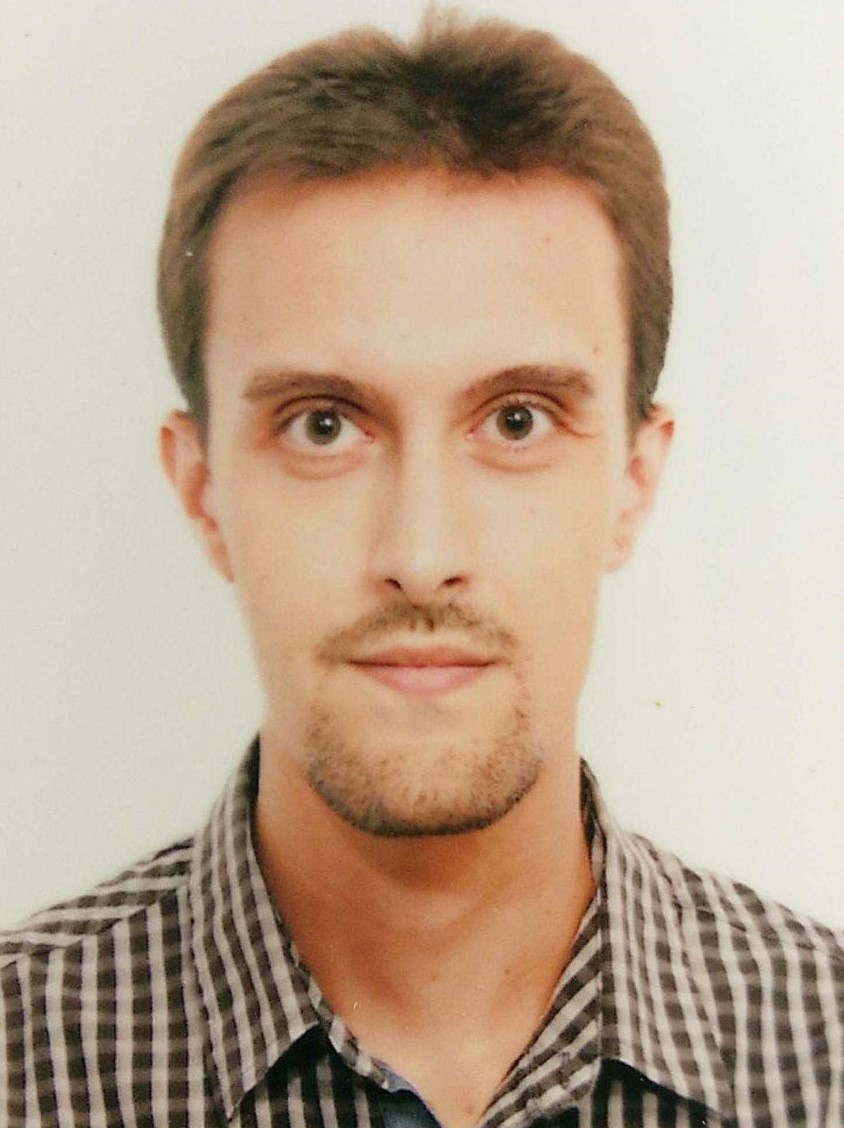}}]{Francisco J. Bald\'{a}n} received an M.Sc. in Data Science and Computer Engineering in 2015 from the University of Granada, Spain. He is currently a Ph.D. student in the Department of Computer Science and Artificial Intelligence, University of Granada, Spain. His research interests include Time Series, Data Mining, Data Science, Big Data, and Cloud Computing.
\end{IEEEbiography}

\begin{IEEEbiography}[{\includegraphics[width=1in,height=1.25in,clip,keepaspectratio]{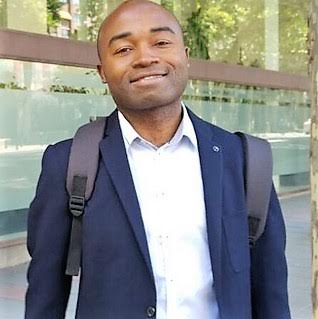}}]{Ghislain Atemezing} 
obtained a Ph.D. in computer science from Telecom ParisTech (France) and is a recognized expert in ontology web language, semantic technologies, and triple store systems. He joined Mondeca (a Semantic software company based in Paris) in 2015. Dr. Atemezing is the Director of Research \& Development at Mondeca. He is  also in charge of promoting, supporting and curating the Linked Open Vocabularies initiative (http://lov.okfn.org/dataset/lov), the reusable linked vocabularies ecosystem shared by ontology experts around the globe. Main work on real-world projects includes ontology modelling, Semantic annotation and reasoning, RDF stores benchmarking, data model validation and fine-tuning.
\end{IEEEbiography}

\begin{IEEEbiography}[{\includegraphics[width=1in,height=1.25in,clip,keepaspectratio]{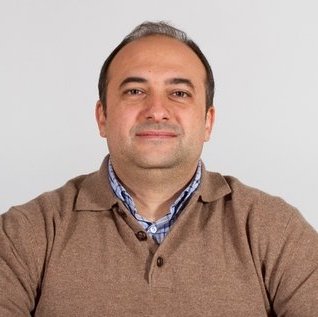}}]{Jose Manuel Benitez-Sanchez}
(M’98) received the M.S. and Ph.D. degrees in Computer Science both 
from the Universidad de Granada, Spain. He is currently an Associate 
Professor at the Department of Computer Science and Artificial 
Intelligence, Universidad de Granada. Dr. Benitez-Sanchez is the head of the 
Distributed Computational Intelligence and Time Series (DiCITS) lab. 
He has been the leading reasearcher in a number of intenational and 
national projects as well as numerous contracts with companies. He has 
co-authored over 70 papers published in international journals with 
contributions ranging from foundations to real world problem 
solutions. His research interests include Cloud Computing, Big Data, 
Data Science, Computational Intelligence, Time Series, Cybersecurity 
and Biometrics.
\end{IEEEbiography}




\end{document}